\PassOptionsToPackage{usenames,dvipsnames}{xcolor}
\documentclass[sigconf,natbib]{acmart}

\usepackage{newfloat}
\usepackage{listings}
\usepackage[utf8]{inputenc} %
\usepackage[T1]{fontenc}    %
\usepackage{url}            %
\usepackage{booktabs}       %
\usepackage{amsfonts}       %
\usepackage{nicefrac}       %
\usepackage{microtype}      %
\usepackage[dvipsnames]{xcolor}         %

\usepackage{todonotes} %

\usepackage[utf8]{inputenc} %
\usepackage{url}            %
\usepackage{booktabs}       %
\usepackage{amsfonts}       %
\usepackage{nicefrac}       %
\usepackage{microtype}      %
\usepackage[dvipsnames]{xcolor}         %
\usepackage{xspace}
\usepackage{parcolumns}

\usepackage{adjustbox}
\usepackage{subcaption}

\usepackage{graphicx}
\usepackage{lipsum}

\usepackage{wrapfig}

\usepackage[utf8]{inputenc}
\usepackage{pgfplots}

\usepackage{hyperref}%

\usepackage{microtype}
\usepackage{graphicx}
\usepackage{booktabs} %

\newcommand{\beq}{\begin{equation}}
\newcommand{\eeq}{\end{equation}}
\newcommand{\bea}{\begin{eqnarray}}
\newcommand{\eea}{\end{eqnarray}}
\newcommand{\ba}{\begin{align*}}
\newcommand{\ea}{\end{align*}}

\usepackage{mathtools} %
\mathtoolsset{showonlyrefs}  

\DeclareMathOperator*{\argmin}{arg\,min}

\AtBeginDocument{%
  \providecommand\BibTeX{{%
    \normalfont B\kern-0.5em{\scshape i\kern-0.25em b}\kern-0.8em\TeX}}}

\setcopyright{usgovmixed}
\copyrightyear{2023}
\acmYear{2023}
\acmDOI{XXXXXXX.XXXXXXX}

\setcopyright{none}

\acmConference[]{}{}{}
\acmBooktitle{} 

\title{Probing the Transition to Dataset-Level Privacy in ML Models Using an Output-Specific and Data-Resolved Privacy Profile
}
  \author{Tyler LeBlond}
  \email{leblond\_tyler@bah.com}
  \affiliation{%
    \institution{Booz Allen Hamilton}
    \streetaddress{304 Sentinel Dr, MD 20701}
    \city{Annapolis Junction}
    \state{Maryland}
    \country{USA}
  }
  \author{Joseph Munoz}
  \email{munoz\_joseph@bah.com}
  \affiliation{%
    \institution{Booz Allen Hamilton}
    \streetaddress{304 Sentinel Dr, MD 20701}
    \city{Annapolis Junction}
    \state{Maryland}
    \country{USA}
  }
  \author{Fred Lu}
  \email{lu\_fred@bah.com}
  \affiliation{%
    \institution{Booz Allen Hamilton}
    \streetaddress{304 Sentinel Dr, MD 20701}
    \city{Annapolis Junction}
    \state{Maryland}
    \country{USA}
  }
  \author{Maya Fuchs}
  \email{fuchs\_maya@bah.com}
  \affiliation{%
    \institution{Booz Allen Hamilton}
    \streetaddress{304 Sentinel Dr, MD 20701}
    \city{Annapolis Junction}
    \state{Maryland}
    \country{USA}
  }
  \author{Elliott Zaresky-Williams}
  \email{zaresky-williams\_elliott@bah.com}
  \affiliation{%
    \institution{Booz Allen Hamilton}
    \streetaddress{304 Sentinel Dr, MD 20701}
    \city{Annapolis Junction}
    \state{Maryland}
    \country{USA}
  }
  \author{Edward Raff}
  \email{raff\_edward@bah.com}
  \affiliation{%
    \institution{Booz Allen Hamilton}
    \streetaddress{304 Sentinel Dr, MD 20701}
    \city{Annapolis Junction}
    \state{Maryland}
    \country{USA}
  }
  \author{Brian Testa} 
  \email{brian.testa.1@us.af.mil}
  \affiliation{
    \institution{Air Force Research Laboratory}
    \streetaddress{26 Electronic Pkwy}
    \city{Rome}
    \state{New York}
    \country{USA}
  }

\renewcommand\footnotetextcopyrightpermission[1]{} 
\fancypagestyle{plain}{%
    \fancyhf{} %
    \fancyfoot[C]{\textbf{\thepage}} %
    \renewcommand{\headrulewidth}{0pt}
    \renewcommand{\footrulewidth}{0pt}}
\begin{document}

\begin{abstract}
Differential privacy (DP) is the prevailing technique for protecting user data in machine learning models. However, deficits to this framework include a lack of clarity for selecting the privacy budget $\epsilon$ and a lack of quantification for the privacy leakage for a particular data row by a particular trained model. 
We make progress toward these limitations and a new perspective by which to visualize DP results by studying a privacy metric that quantifies the extent to which a model trained on a dataset using a DP mechanism is ``covered" by each of the distributions resulting from training on neighboring datasets. We connect this coverage metric to what has been established in the literature and use it to rank the privacy of individual samples from the training set in what we call a privacy profile. We additionally show that the privacy profile can be used to probe an observed transition to indistinguishability that takes place in the neighboring distributions as $\epsilon$ decreases, which we suggest is a tool that can enable the selection of $\epsilon$ by the ML practitioner wishing to make use of DP. 
\end{abstract}

\maketitle

\section{Introduction}\label{sec:intro}
With the privacy of individuals' data becoming of greater concern to the Machine Learning (ML) community, the theoretical guarantees of \textit{differential privacy} have become highly attractive. Differential privacy (DP) is the prevailing mathematical framework for ensuring the privacy of data used by randomized algorithms. Its application to the training process of ML algorithms is well-studied and has been implemented for general use in packages such as IBM's Diffprivlib \citep{holohan2019diffprivlib}. The definition of $\epsilon$-DP \citep{dwork2014algorithmic} can be written in the form of a bound on the maximum \textit{privacy loss} as \beq \label{privloss} \epsilon \geq \log{\frac{\text{Pr}\left[\mathcal{A}(x)\in S\right]}{\text{Pr}\left[\mathcal{A}(y)\in S\right]}} \equiv l(x,y,S),\eeq where $\mathcal{A}(x)$ is a randomized algorithm acting on a dataset $x$, a so-called neighboring dataset $y$ has one row added to or removed  from $x$, $S$ is a subset of the output space of $\mathcal{A}$, and $l(x,y,S)$ is the privacy loss. The maximum privacy loss, which is bounded by $\epsilon$, depends on the worst-case combination of dataset pair $x, y$ and output subspace $S$. In the context of DP Machine Learning, it is commonly understood that Eq.~\eqref{privloss} sets a constraint on an attacker's ability to discern whether an individual's data is present in the training set of a given model \citep{yeom2018privacy}.

Currently, using differential privacy for ML applications involves training models with an $\epsilon$ value that sufficiently constrains the privacy loss without significantly reducing output model accuracy \citep{abadi2016deep}. The selection of an appropriate $\epsilon$ is difficult because the guarantee of Eq.~\eqref{privloss} applies to \textit{all possible} training datasets and \textit{all possible} subsets of the mechanism output space (here, the model parameters). Because of this, it is not surprising that in practical applications ridiculously high values of $\epsilon$ give much stronger effective privacy than expected \citep{rahman2018membership, giraldo2020adversarial}. Therefore, the practicality of Eq.~\eqref{privloss} in the context of ML is limited for two reasons. First, the goal is to train a model on a particular dataset that rarely (together with its neighbors) maximizes the right-hand side. Second, one usually intends to produce a single model that is subsequently used, and thus the global privacy of the model-training mechanism is irrelevant; a model-specific privacy definition is needed. Recent works have proposed more practical versions of Eq.~\eqref{privloss} that account for these concerns~\citep{soria2017individual, wang2019per, redberg2021privately}. Here, we introduce a conceptual framework that enables model and data-specific privacy quantification for models that are trained by $\epsilon$-DP mechanisms in the sense of Eq~\eqref{privloss}. Insofar as we apply the \textit{ex-post} per-instance privacy loss, our work is complementary to the recently introduced model and data-specific notion of DP known as \textit{ex-post} per-instance DP (or \textit{ex-post} pDP)~\citep{redberg2021privately}.

\begin{figure*}
    \centering
    \includegraphics[width=\textwidth]{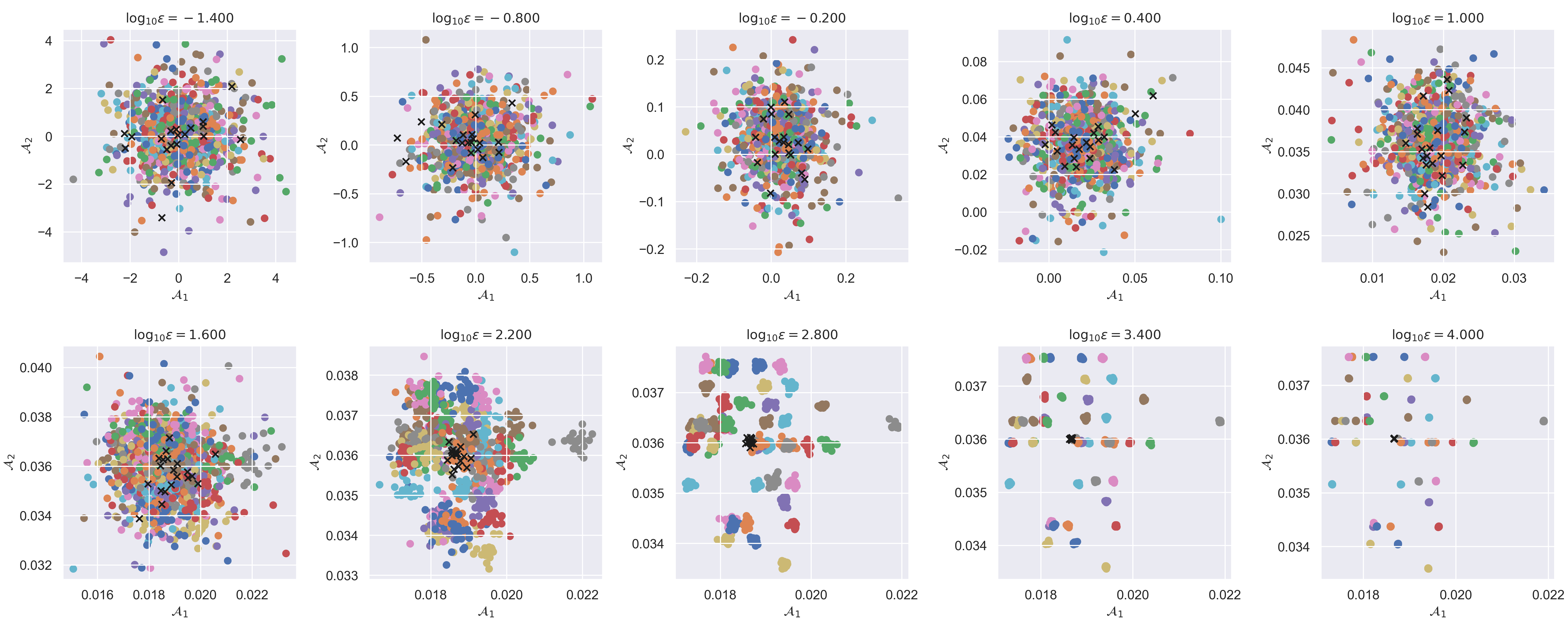}
    \caption{Samples from Diffprivlib's logistic regression on 100 rows of a 2-feature version of the Adults dataset (black x's). Colored circles are the same results on 50 neighboring datasets. $\epsilon$ is varied over the subplots (see subplot titles), demonstrating the increase of variance in the output space for decreasing $\epsilon$. This shows that once $\epsilon$ is sufficiently large or small, no change in privacy occurs for small changes is $\epsilon$. Our framework will show how to visualize this in a way more useful to a practitioner than current opaque ``test and checl''. Importantly, this also lets the practioner understand which part of a dataset is hindering/lacking privacy (e.g., the grey population in the above plots). }
    \label{fig:coverage}
\end{figure*}

We note that an interpretation of Eq.~\eqref{privloss} for ML applications is that distributions of models trained on neighboring datasets should be indistinguishable if the training mechanism is to be considered private, where $\epsilon$ controls the degree of distinguishability. In this article, we attempt to quantify this indistinguishability starting with the \textit{coverage}, which is the probability that the mechanism of interest $\mathcal{A}$ on a dataset of interest $y$ produces a particular model point $M$ (expressed as a vector in the model parameter space $\theta$). The coverage is written as $C(y, M) = f_y(M)$, where $f_y(M)$ is the pdf underlying $\mathcal{A}(y)$ evaluated at model point $M$.\footnote{For cases in which the pdf underlying $\mathcal{A}(y)$ is unknown, the coverage is calculated empirically using the fraction of samples from $\mathcal{A}(y)$ that lie within a cutoff radius $c$ of $M$. In that case, the coverage is written as $C(y, M) = \mathbb{P}(|\mathcal{A}(y)-M|<c)$, where $|\mathcal{A}(y)-M|$ is the $L_2$ distance between samples from $\mathcal{A}(y)$ and the model point $M$.} A connection can be made to Eq.~\eqref{privloss}, as a form of privacy loss can now be expressed as $l(x,y,M)=-\log{\frac{C(y,M)}{C(x,M)}}$. This quantity represents the log of the ratio of probabilities that $\mathcal{A}(x)$ vs. $\mathcal{A}(y)$ produce $M$: if $C(y, M)$ is very different from $C(x, M)$, then the row missing from $y$ is vulnerable in $M$. $l(x, y, M)$ is thus an intuitive privacy loss that is constrained to the relevant scenario for ML applications and resolves the aforementioned difficulties. 

The \textit{ex-post} pDP is a rigorous framework that establishes bounds on $|l(x,y,M)|$ while providing privacy guarantees~\citep{redberg2021privately}. We take a separate but related perspective, envisioning the same privacy loss as part of the ML practitioner's toolkit to gauge the practical privacy of their output model that was trained using $\epsilon$-DP. To accomplish this, we apply $|l(x,y,M)|$ to the definition of a privacy profile for training samples in individual models and discuss the use of this profile as a probe of the transition to dataset-level privacy for typical models. Specifically, we have demonstrated that the privacy profile averaged across different model samples $M\in \mathcal{A}(x)$ converges as $\epsilon$ decreases to a static function, which we claim implies that there is no benefit to decreasing $\epsilon$ beneath a certain threshold value. To strengthen this claim, we have studied the sample mean of $|l(x,y,M)|$ over many $M$ for different neighboring datasets $y$. The expected value of $|l(x,y,M)|$ has been noted elsewhere to be the Kullback-Leibler (KL) divergence of $\mathcal{A}(x)$ and $\mathcal{A}(y)$~\citep{dwork2016concentrated,cuff2016differential}. We found that this quantity hits a plateau at a critical value of $\epsilon$ that scales inversely with the distance between distribution centers $A(x)$ and $A(y)$. For all neighboring datasets, the curves plateau at $\epsilon$ values above the aforementioned threshold value, implying that \textit{all} samples are protected beneath that threshold.

The rest of our paper, and each section's main contribution, is organized as follows. In Sec.~\ref{sec:motivation}, we motivate the definition of coverage and the privacy profile through a qualitative discussion of results obtained from IBM's Diffprivlib. In short, we set the stage with the phenomena that we seek to understand using these metrics. In Sec.~\ref{sec:setup} we provide the problem setup, including various technical definitions and the details of how we implement $\epsilon$-DP. In Sec.~\ref{sec:coverage_formalism}, we explain the coverage formalism and analytically extract the relevant parameters and variables that determine privacy loss given our problem setup. In Sec.~\ref{sec:numerical_results}, we discuss numerical calculations of the coverage for both individual and typical model points, demonstrating empirically that the typical privacy profile can be used to describe three regimes in $\epsilon$ in which (a) all effective privacy has been obtained, (b) there is yet no privacy, and (c) the transition between these. We finish with related works in Sec.~\ref{sec:related_work} followed by our conclusions in Sec.~\ref{sec:conclusion}.
Lastly, Appendix~\ref{sec:motivation2} shows that the qualitative phenomena noted in Sec.~\ref{sec:motivation} are equally apparent using this method. 

\section{Coverage Motivation} \label{sec:motivation}

To motivate our discussion, the first mechanism $\mathcal{A}$ that we consider is diffprivlib's DP logistic regression~\citep{holohan2019diffprivlib}, which employs objective perturbation with Laplace noise to implement differential privacy~\citep{chaudhuri2011differentially}. We choose this as an example of an `out of the box' implementation of differential privacy that somebody looking to take advantage of DP in ML classification might use. Figure~\ref{fig:coverage} shows the output of 20 samples from the \textit{base} distribution $\mathcal{A}(x)$ (black x's), where $x$ is the first 100 rows of a 2-feature version of the Adults dataset\footnote{We select the first and fifth columns of the Adults dataset and normalize them as follows: first, we scale each feature so that it has a mean of zero and a variance of 1. Then, we normalize data rows by the largest row magnitude so that $|\mathbf{x}_i|\leq 1$ everywhere.}. We also show 20 samples from 50 different \textit{neighboring} distributions $\mathcal{A}(y_i)$ (colored circles), where $\{y_i\}$ in this case are the 100 single-row-removed neighboring datasets to $x$. To visualize the effects of DP in the output space, we vary $\log_{10}\epsilon$ throughout the subplots (see subplot titles for details), making sure to include a range of values that captures an apparent transition between two poles of output behavior. We observe that at high values of $\log_{10}\epsilon$, $\mathcal{A}(x)$ and all $\mathcal{A}(y_i)$ are clearly distinguishable and well-localized in the output space (i.e., no privacy). At low $\log_{10}\epsilon$, the same distributions are completely indistinguishable from one another by sight (i.e., private but uninformative). There is a narrow intermediate regime in $\log_{10}\epsilon$ where the behavior is mixed, which reveals the practical range of usable privacy vs accuracy trade-off. A goal of this paper is to capture this transition mathematically.

It is clear from Fig.~\ref{fig:coverage} that there is a competition of scales in the output behavior. First, there is the scale set by the scale parameter in the Laplace distribution (inversely related to $\epsilon$) that dictates the spread of output model points in parameter space. Second, there are the scales set by the distances between neighboring distribution centers and the base distribution center. As $\log_{10}\epsilon$ decreases from high values, the noise added in $\mathcal{A}$ improves the coverage of sampled $M \in \mathcal{A}(x)$ by the neighboring distributions $\mathcal{A}(y_i)$, which can seen through the apparent increase in distribution overlap. Beneath a $\log_{10}\epsilon$ threshold, however, added variance does not improve this coverage because the spread of the distributions has already increased beyond the scales set by the distribution centers. This global transition to indistinguishability (which we will refer to as dataset-level privacy) begins at the level of individual models, which each distinguish differently between the neighboring distributions. The coverage, formalized below, provides an intuitive model-level evaluation of privacy for each row of the training set in the form of a \textit{privacy profile}. We then probe the transition to indistinguishability using the intuition that it can be seen through the privacy profile of typical models.

\section{Problem Setup} \label{sec:setup}
\subsection{Empirical Risk Minimization}
 We suppose that our mechanism $A$ is the training algorithm of a binary linear classifier $\mathbf{f}: \mathcal{X}\to\mathcal{Y}$ that takes input from the data space $\mathcal{X}=\mathbb{R}^d$ to the class label space $\mathcal{Y}=\{-1,+1\}$. $A$ receives input in the form of an $n$-sample training dataset $\mathcal{D} = \{(\mathbf{x}_i, y_i)\in\mathcal{X}\times\mathcal{Y}:i=1,2,...,n\}$ and outputs a vector of model parameters $\mathbf{f}$. Following \cite{chaudhuri2011differentially}, we assume that $\mathcal{X}$ is the unit ball such that $|\mathbf{x}_i|\leq 1$.

Following the assumptions of ~\citep{chaudhuri2011differentially, zhou2020tighter}, we consider the case of $L_2$ regularized loss functions \beq J(\mathbf{f}, \mathcal{D}) = \frac{1}{n}\sum_{i=1}^n l(\mathbf{f}(\mathbf{x}_i),y_i) + \frac{\Lambda}{2} |\mathbf{f}|^2,\eeq where $\mathbf{f}(\mathbf{x}) = \mathbf{f}^\top\mathbf{x}$ and $\Lambda$ is the regularization constant, which we set to be $\Lambda = 1$ unless otherwise stated. In the present study, all numerical examples use the logistic regression loss function \beq l(\mathbf{f}(\mathbf{x}_i), y_i) = \ln{(1+e^{-y_i \mathbf{f}(\mathbf{x}_i)})}, \eeq though our results are not dependent on the specific loss. The trained model results from minimizing $J(\mathbf{f},\mathcal{D})$ with respect to the model parameters $\mathbf{f}$, as in \beq A(\mathcal{D}) = \argmin_{\mathbf{f}} J(\mathbf{f},\mathcal{D}). \eeq

We consider datasets $\mathcal{D}$ and $\mathcal{D}'$, where $\mathcal{D}'$ is equivalent to $\mathcal{D}$ with a single row removed and thus has $n-1$ entries. We call $A(\mathcal{D})$ the \textit{base} model point and all $n$ possible $A(\mathcal{D}')$ are called \textit{neighboring} model points. From now, we will switch to the notation $x$ for the base dataset and $\{y_i\}$ is the set of $n$ row-removed neighboring datasets (we used $\mathcal{D}$, $\mathcal{D}'$ above so as to avoid confusion with the notation for data row and target variable). 

\subsection{DP Implementation}  \label{sec:dp_implementation}
To implement $\epsilon$-DP, we perturb base and neighboring model points with Laplace noise as in \cite{chaudhuri2011differentially}. We consider $\mathcal{A}(x) = A(x) + \mathbf{b}$, where $v(\mathbf{b})=\frac{\beta}{2} e^{-\beta|\mathbf{b}|}$ is the distribution underlying the random vector $\mathbf{b}$ and $\beta = \frac{n\Lambda\epsilon}{2}$. In what follows, we will denote the probability distribution function underlying $\mathcal{A}(x)$ as $f_x(M)$, where $M$ is a point in model parameter space (which we generically call a model point).

While in this paper we choose to focus on the simpler output perturbation method to implement DP, we note that objective perturbation method, proposed in \cite{chaudhuri2011differentially}, is the more popular technique. We stick to output perturbation here for purposes of demonstration, since coverage calculations are simpler when the distribution of model points in the mechanism output space is known \textit{a priori}. In complementary fashion, the theoretical analysis in \cite{redberg2021privately} has shown how to efficiently obtain \textit{ex-post} pDP privacy losses for objective perturbation in linear models. Given the intuition suggested by the results in Fig.~\ref{fig:coverage}, which was obtained using objective perturbation, we expect the key results of the present work to generalize as long as distributions are sufficiently well-behaved in the mechanism output space. Given an efficient strategy for calculating the privacy loss, our methods for generating privacy profiles and using them to probe the aforementioned transition to indistinguishability should apply.

\subsection{Application to Unlearning}

As our method and discussion will note, we use machine unlearning to implement and explore coverage with linear models. The linear unlearning methods are currently the most effective, and so form the basis of our study. In particular, they can be done with near-exact results in bounded time, as we demonstrate in \autoref{sec:bounds}. As unlearning methods are improving for kernel, decision tree, and deep neural networks \cite{Bourtoule,Brophy_Lowd_2021,ApproxData,Sivan_Gabel_Schuster_2021,Zhou2021}, they will become more viable for using coverage in the future. 

In the short term, our method provides a new approach to studying how to infer a meaningful value of $\epsilon$. In particular, we can deduce a range of meaningful values of $\epsilon \in [\mathit{low}, \mathit{high}]$, such that any value of $\epsilon$ outside this range is not worth testing. Values of $\epsilon \geq \mathit{high}$ will all be equally unprotected, and such larger values provide no change in privacy. Similarly, any value of $\epsilon \leq \mathit{low}$ will be one that has obtained the maximum effective level of privacy obtainable given a specific dataset, and so no value lower is worth consideration. 

\section{Coverage Formalism} \label{sec:coverage_formalism}
\subsection{Coverage Definition} 
In this section, we present a general framework for analyzing the model and data-specific privacy of machine learning models that were trained with differential privacy in mind. In order to do this, we quantify the probability that an $\epsilon$-DP mechanism $\mathcal{A}$ produces model points nearby a target model $M$ in the parameter space. 

The \textit{coverage} is the probability that a randomized mechanism $\mathcal{A}$ over a dataset $y$ produces output within a radius $c$ of a target output model point $M$. In the machine learning context, we take $M$ as a vector of model parameters that is sampled from $\mathcal{A}(x)$, where $\mathcal{A}(x)$ is a random vector corresponding with the output of a random mechanism over the \textit{base} dataset $x$. The purpose of this setup is that one can think of $M$ as a trained machine learning model generated using a DP ML algorithm (and that one wishes to analyze the privacy of). We define the coverage of that point by the same mechanism over \textit{neighboring} dataset $y$ as \beq \label{eq:cov1} C(y, M) = \mathbb{P}(|\mathcal{A}(y) - M| < c), \eeq where $c$ is an arbitrary $\mathcal{O}(1)$ cutoff radius. Note that $C(x, M)$ is a special case that we call the \textit{self-coverage} because it represents the analogous probability for the base distribution.

In this paper we will discuss the properties of the coverage and self-coverage, but will be most interested in the ratio of these, which we formulate as a privacy loss as in \beq \label{eq:ploss} l(x, y, M) = -\log\left[C(y,M)/C(x,M)\right]. \eeq 

We note that, to compute $C(y,M)$ in general, one has to compute an integral of the pdf underlying $\mathcal{A}(y)$ over the sphere centered at $M$ of radius $c$. The effort required for this computation depends significantly on the assumptions involved. However, with \textit{a priori} knowledge of the distribution of model points in the output space of $\mathcal{A}(y)$, and in the limit of $c\to 0$, the coverage becomes a simple evaluation of the pdf at the target point $M$. In other words, we can let \beq \label{eq:cov2} C(y, M) = f_y(M) \eeq where $f_y(M)$ is the pdf underlying $\mathcal{A}(y)$ evaluated at model point $M$. In this case, we have 
\beq \label{eq:ploss2} l(x, y, M) = -\log\left[f_y(M)/f_x(M)\right]. \eeq At this point, we remind the reader that this quantity has been introduced and studied elsewhere in a slightly different context~\citep{redberg2021privately}. While others have focused on bounding this quantity in order to present a reformulation of DP known as \textit{ex-post} pDP, here we present examples of its utility to the ML practitioner who wishes to use $\epsilon$-DP to protect their dataset.

As mentioned in Sec.~\ref{sec:dp_implementation}, for ease of computation, in the remainder of this paper we will assume Laplace distributed noise in the output space as in ~\citep{chaudhuri2011differentially}. This is consistent with the use of an output perturbation technique to implement differential privacy. Preliminary results have suggested that our key findings apply to Gaussian-distributed noise as well, indicating the generality of our results.

\subsection{Laplace-distributed Noise}\label{sec:coverage_formalism2}
In the case of Laplace-distributed noise, Eq.~\eqref{eq:ploss2} becomes \beq \label{eq:ploss3} l(x, y, M) = -\log\left[e^{-\beta |A(y)-M|}/e^{-\beta |A(x)-M|}\right], \eeq where it has been assumed that the Laplace distributions share an $\epsilon$-dependent scale parameter $\beta$. Eq.~\eqref{eq:ploss3} simplifies to $l(x, y, M) = \beta\left[|A(y)-M| - |A(x)-M|\right]$. Thus, for a given value of $\epsilon$, the coverage implies that the more different the distances between the distribution centers and $M$ are, the higher the privacy loss, and thus the more vulnerable is the relevant row of data (the one different between $x$ and $y$) for model point $M$. With fixed locations of $M$ and distribution centers, the privacy loss simply scales linearly with $\beta$.

Because we consider true privacy for a $y$, $M$ pair to be the case where $l(x,y,M) = 0$, we consider the absolute value of the privacy loss to be the most relevant quantity. Defining $d(x,y,M) = \left||A(y)-M| - |A(x)-M|\right|$, we write $|l(x,y,M)| = \beta d(x,y,M)$. This form succinctly expresses the competition of distance scales between the scale of the Laplace noise and a distance scale $d$, formalizing our intuition from Sec.~\ref{sec:motivation}. For a given $M$, it is $d_{\mathrm{max}} \equiv d(x,y_{\mathrm{max}},M) \equiv \max_{y} d(x,y,M)$ that sets the maximum privacy loss and the most vulnerable data row is the difference between $y_{\mathrm{max}}$ and $x$. We emphasize that the vulnerability ranking for $M$ is static with respect to $\beta$, suggesting that such a ranking is also well-defined in the $\beta \to \infty$ limit where $M$ was not trained with differential privacy in mind.  This has implications for quantifying the relative privacy of rows within the training set of ML models that were not trained to be private.

\subsection{Privacy Profile} \label{sec:privacyprofile}
Using the coverage-motivated definition of privacy loss, we further define a \textit{privacy profile} for general model points $M$. We define the privacy profile for a given $M$ to be the ranked absolute values of privacy loss $|l(x,y_{i_M},M)|$, where $\{i_M\}=\mathrm{argsort}\left[|l(x,y_i,M)|\right]$ is the ordered set of indices for neighboring datasets to $x$. The privacy profile for $M$ not only gives us the ranked list of vulnerable rows in $x$, but additionally tells us how privacy losses vary throughout the spectrum of neighboring datasets when represented as a function of ranking. Later, we will discuss what the privacy profile for typical $M$ can tell us about the transition to indistinguishability mentioned in Sec.~\ref{sec:motivation}.

\subsection{Coverage for Typical Models}
While coverage is well-defined for any point $M$ in the output space of $\mathcal{A}$, most of these points are irrelevant because they are not likely to be sampled when drawing from $\mathcal{A}(x)$. In order to study the coverage properties of models that are \textit{typically} sampled from the base distribution $\mathcal{A}(x)$, we study the expected value of coverage-relevant quantities [such as Eq.~\eqref{eq:ploss3}] over this distribution. Our ultimate goal is to make statements about the privacy loss behavior of typical $M \in \mathcal{A}(x)$ as a function of $\beta$. In order to characterize privacy loss for typical $M$ as a function of $\beta$, we need $\mathbb{E}[l(x,y,M)] = - \mathbb{E}\left[\log{\left[\frac{C(y,M)}{C(x,M)}\right]}\right] = \mathbb{E}[\log{C(y,M)}] - \mathbb{E}[\log{C(x,M)}]$. Unfortunately, while $\mathbb{E}[C(x,M)]$ (and $\mathbb{E}[\log{C(x,M)}]$) are solvable in closed form for the case studied here, $\mathbb{E}[C(y,M)]$ and $\mathbb{E}[\log{C(y,M)}]$ are not. Thus, in order to understand $\mathbb{E}[l(x,y,M)]$,\footnote{We note that this quantity is the Kullback-Leibler (KL) divergence of $\mathcal{A}(x)$ and $\mathcal{A}(y)$~\citep{cuff2016differential,dwork2016concentrated}.} we will resort to numerical averaging of Eq.~\eqref{eq:ploss3} over many samples of $M$. We study the typical privacy profile in the same way.

\section{Numerical Results} \label{sec:numerical_results}

\subsection{Approximate Location of Neighboring Model Point}\label{sec:bounds}
Our starting point is the bounds for $A(y)$ that were derived in \cite{okumura2015quick} and ~\cite{zhou2020tighter}. The bounds from these papers constrain $A(y)$ to a sphere and to the half-space divided by a plane, respectively. These works considered cases where the bound showed removing a row of the dataset could not change the classification for a point, allowing for fast cross-validation with minimal retraining for only the points that could not be pruned. 

\begin{figure}[!h]
    \centering
    \begin{adjustbox}{max width=\columnwidth}
    \includegraphics{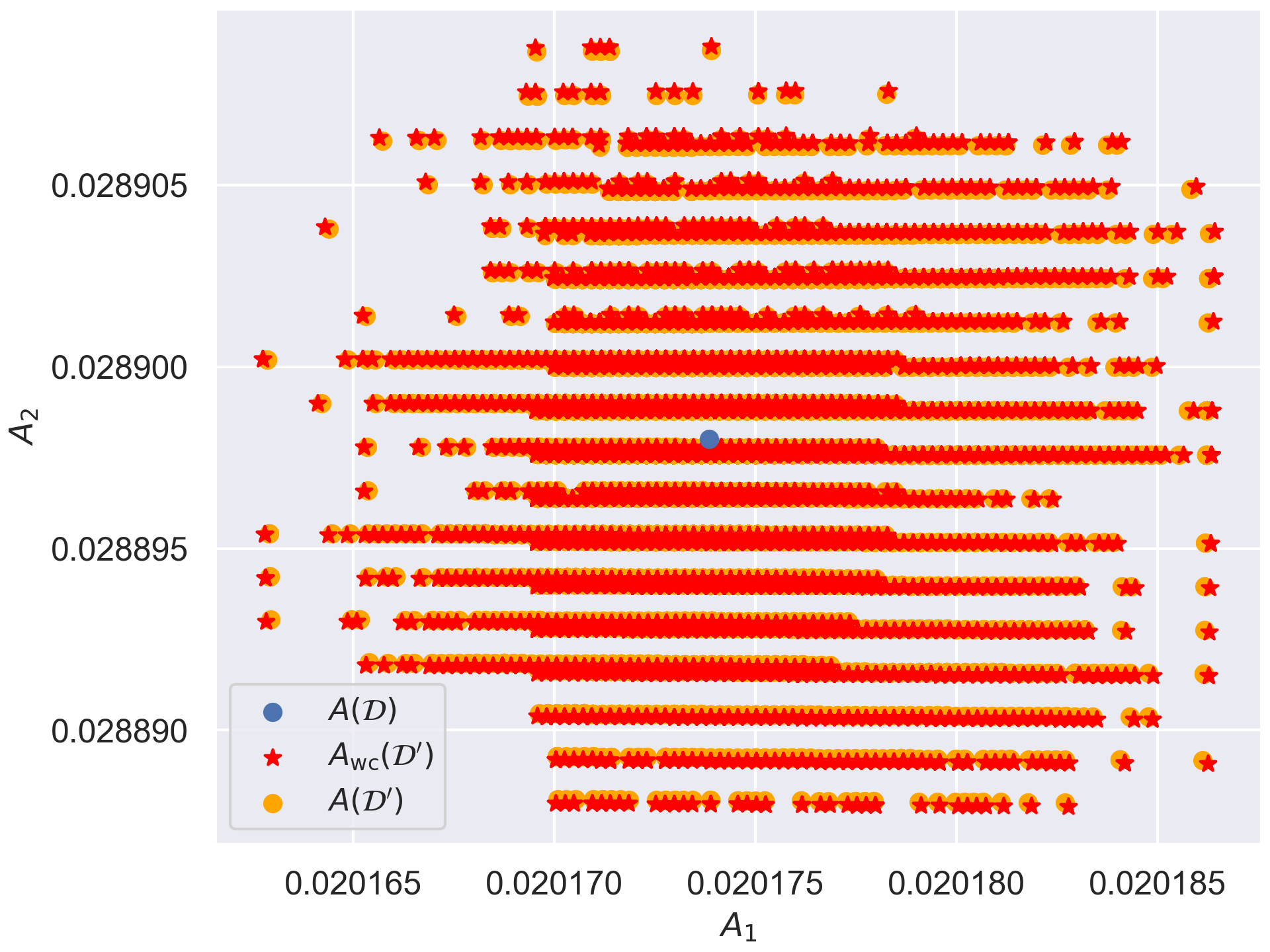}
    \end{adjustbox}
    \caption{Full (two-column) adults dataset comparison of $A_{\mathrm{wc}}(y)$ (our projected worst cases) and $A(y)$ (exhaustive model retraining) show that our approach results in accurate estimates of the models of neighbor datasets, and thus accurate coverage estimates.}
    \label{fig:wc_vs_exact}
\end{figure}

In our work we require the exact solution of the hyper-plane for all neighboring datasets $\{y_i\}$, but can not pay the penalty of multiple training. Here, we take the original solution for the base dataset and project it to the farthest point in the convex hull defined by the union of the bounds by \citep{okumura2015quick,zhou2020tighter}. This is demonstrated empirically to provide a good approximation to the exact solution. One can see this visually in Fig.~\ref{fig:wc_vs_exact} for a two-dimensional feature space, and we have confirmed numerical closeness in higher dimensions. For example, using 6 features of the full Adults dataset, we found that $\max{\left[|A_{\mathrm{wc}}(y)-A(y)|/|A(y)-A(x)|\right]} < 2\times 10^{-3}$. In other words, the deviation of our projected worst-case model point from the one obtained by exhaustive model retraining is less than $0.2\%$ of the distance between the exact base and neighboring model points. That small deviation is negligible on the length scales relevant to the problem. We provide a more rigorous description of $A_{\mathrm{wc}}(y)$ in \autoref{sec:boundsjust}. 

\subsection{Worst-Case Point Analysis and Justification}\label{sec:boundsjust}
In Sec.~\ref{sec:bounds}, we presented results showing empirically that the so-called worst-case neighboring model point within certain theoretical bounds is approximately equal to the exact result found through retraining. Here, we will expand on our study of the location of neighboring model points and motivate our decision to use the worst-case point.

First, we re-iterate our problem setup, describe the spherical bound introduced in Ref.~\citep{okumura2015quick}, and reduce it to our particular case. Continuing with the notation of the main article, we let $A(y_i)$ be the exact model point on the $i$th row-removed neighboring dataset $y_i$. It is the vector of model parameters that minimizes the $L_2$ regularized loss function $J(\mathbf{f},y_i)$. In this section, we seek to locate this point to a good approximation without re-training the model from scratch. In order to do so, we leverage both analytical bounds from the literature and other insights that we have discovered empirically.

Given a general dataset $\mathcal{D} = \{(\mathbf{x}_i, y_i)\in\mathcal{X}\times\mathcal{Y}:i=1,2,...,n\}$, where $\mathcal{X}=\mathbb{R}^d$ and $\mathcal{Y}=\{-1,+1\}$ are respectively the data space and the class label space, the regularized loss function is defined as follows\footnote{From here to the end of paragraph is copied directly from the main article. Note that, while throughout this writeup $y_i$ refers to the $i$th neighboring dataset to $x$ (the base dataset), in this paragraph $y_i$ is the class label of the $i$th data sample.} \beq J(\mathbf{f}, \mathcal{D}) = \frac{1}{n}\sum_{i=1}^n l(\mathbf{f}(\mathbf{x}_i),y_i) + \frac{\Lambda}{2} |\mathbf{f}|^2,\eeq where $\mathbf{f}(\mathbf{x}) = \mathbf{f}^\top\mathbf{x}$ and $\Lambda$ is the regularization constant. We set $\Lambda = 1$ unless otherwise stated. In the present study, all numerical examples use the logistic regression loss function \beq l(\mathbf{f}(\mathbf{x}_i), y_i) = \ln{(1+e^{-y_i \mathbf{f}(\mathbf{x})})}, \eeq though our results are not dependent on the specific loss. The trained model results from minimizing $J(\mathbf{f},\mathcal{D})$ with respect to the model parameters $\mathbf{f}$, as in \beq A(\mathcal{D}) = \argmin_{\mathbf{f}} J(\mathbf{f},\mathcal{D}). \eeq From now, we will switch back to the notation where $x$ denotes the base dataset and $\{y_i\}$ denotes the set of $n$ row-removed neighboring datasets.

The spherical region $S(\mathbf{R}, r)$ (first introduced in Ref.~\citep{okumura2015quick}) that constrains $A(y_i)$ can be written as \beq \label{eq:sphere} S(\mathbf{R}, r) \equiv \{A(y_i): |A(y_i)-\mathbf{R}|\leq r\}, \eeq where (in general)\footnote{Here, $n_0$ and $n_1$ are respectively the number of samples in the original and modified datasets, and $n_{\mathcal{A}}$ and $n_{\mathcal{S}}$ are respectively the number of samples added and removed from the original dataset to create the modified one.} \beq \label{eq:q} \mathbf{R} = \frac{n_0 + n_1}{2n_1} A(x) - \frac{n_{\mathcal{A}} + n_{\mathcal{S}}}{2\Lambda n_1}\Delta s,\eeq \beq \label{eq:r} r = \left|\frac{n_{\mathcal{A}}-n_{\mathcal{S}}}{2n_1}A(x) + \frac{n_{\mathcal{A}} + n_{\mathcal{S}}}{2\Lambda n_1}\Delta s\right|,\eeq and \beq \label{eq:ds} \Delta s = \frac{1}{n_{\mathcal{A}}+n_{\mathcal{S}}}\left(\sum_{i\in\mathcal{A}}\nabla l_i[A(x)] - \sum_{i\in\mathcal{S}}\nabla l_i[A(x)] \right) \eeq

In our special case of single row-removed neighboring datasets, we let $n_0 = n$, $n_1 = n-1$, $n_{\mathcal{A}} = 0$, and $n_{\mathcal{S}} = 1$. This reduces \eqref{eq:ds} to $\Delta s = -\nabla l_R[A(x)]$, where $\nabla l_R[A(x)]$ is the gradient of the loss of the base model point on the removed row. Then, \eqref{eq:q} becomes \beq \label{eq:q2} \mathbf{R} = \left(1+\frac{1}{2(n-1)}\right) A(x) + \frac{1}{2\Lambda(n-1)}\nabla l_R[A(x)]\eeq and \eqref{eq:r} becomes \beq \label{eq:r2} r = \frac{1}{2(n-1)}\left|A(x) + \frac{1}{\Lambda}\nabla l_R[A(x)]\right|.\eeq

A few comments about the asymptotic behavior of $r$, which represents the tightness of the spherical bound, are in order. First, from \eqref{eq:r}, one can see that $r \to 0$ as $n_1 \to \infty$ so long as $n_{\mathcal{A}}$ and $n_{\mathcal{S}}$ are both $\mathcal{O}(1)$, which is the case in \eqref{eq:r2}. We additionally note that $r \to \infty$ as $\Lambda \to 0$ (all else held constant) and $r \to \frac{|n_{\mathcal{A}} - n_{\mathcal{S}}|}{2n_1}|A(x)|$ as $\Lambda \to \infty$. We have assumed that $\Delta s$ is finite. Evidently, the spherical bound implies an exact location of $A(y_i)$ in the limit of large $n$, and the regularization parameter can also be used to tune the tightness of the bounds toward its optimal value (for a given $n$) at high $\Lambda$. Because of this, we might be tempted to believe that if our dataset is large, we can safely make the approximation that $A(y_i)\simeq \mathbf{R}$. This would be helpful to us if it were not for the following property of the bounds. In our special case,

\bea 
\frac{r}{|\mathbf{R} - A(x)|} &=& \frac{\frac{1}{2(n-1)}\left|A(x) + \frac{1}{\Lambda}\nabla l_R[A(x)]\right|}{\left|\frac{1}{2(n-1)} A(x) + \frac{1}{2\Lambda(n-1)}\nabla l_R[A(x)]\right|}\\
&=& \frac{\left|A(x) + \frac{1}{\Lambda}\nabla l_R[A(x)]\right|}{\left|A(x) + \frac{1}{\Lambda}\nabla l_R[A(x)]\right|}\\
&=& 1.
\eea

Thus, $r = |\mathbf{R} - A(x)|$, which implies that $A(x)$ is always located on the surface of the sphere that bounds $A(y_i)$. We note that this result is true in general, and can be easily proved using the fact that $n_0 - n_1 = n_{\mathcal{S}} - n_{\mathcal{A}}$. 

At this point, we can define $A_{\mathrm{wc}}(y_i)$, the so-called \textit{worst-case} neighboring model point located directly across the sphere from $A(x)$, as \beq A_{\mathrm{wc}}(y_i) = A(x) + 2[\mathbf{R} - A(x)] = 2\mathbf{R} - A(x).\eeq The farthest distance possible between the neighboring and base model points is $|A(y_i) - A(x)|_{\mathrm{max}} = |A_{\mathrm{wc}}(y_i) - A(x)| = 2|\mathbf{R}-A(x)| = 2r \sim 1/n$. This means that $|A(y_i) - A(x)|$ approaches zero at least as quickly with $n$ as $r$ does. In conclusion, while the bound tightness scales nicely with $n$, increasing $n$ does not help us better locate $A(y_i)$ with respect to $A(x)$. 

In order for a bound on $A(y_i)$ to be adequate for the purposes of calculating coverage, it needs to be tight enough that the error in $A_{\mathrm{guess}}(y_i)$ is significantly smaller than the distance between $A_{\mathrm{guess}}(y_i)$ and $A(x)$. It is necessary to resolve between $A(y_i)$ and $A(x)$ for all $y_i$ in order to accurately calculate coverage and generate privacy profiles. The naive use of the spherical bound through letting, say, $A(y_i) \simeq \mathbf{R}_{y_i}$ does not yield good resolution because then $r/|A(y_i)_{\mathrm{guess}} - A(x)|=1$.

A tighter constraint on the location of $A(y_i)$ has been published in Ref.~\citep{zhou2020tighter}. In that article, a plane dividing the possible parameter space in half is proposed. This plane always intersects the sphere from Ref.~\citep{okumura2015quick}, meaning that the bounds on $A(y_i)$ are always improved with respect to the sphere. Still, these bounds do not improve our ability to locate $A(y_i)$ for the purpose of computing coverage because they do not change the location of the worst-case point: in fact, it can be easily shown that the worst-case point defined above always intersects the plane. 

\begin{figure}[!h]
    \centering
    \includegraphics[width=\columnwidth]{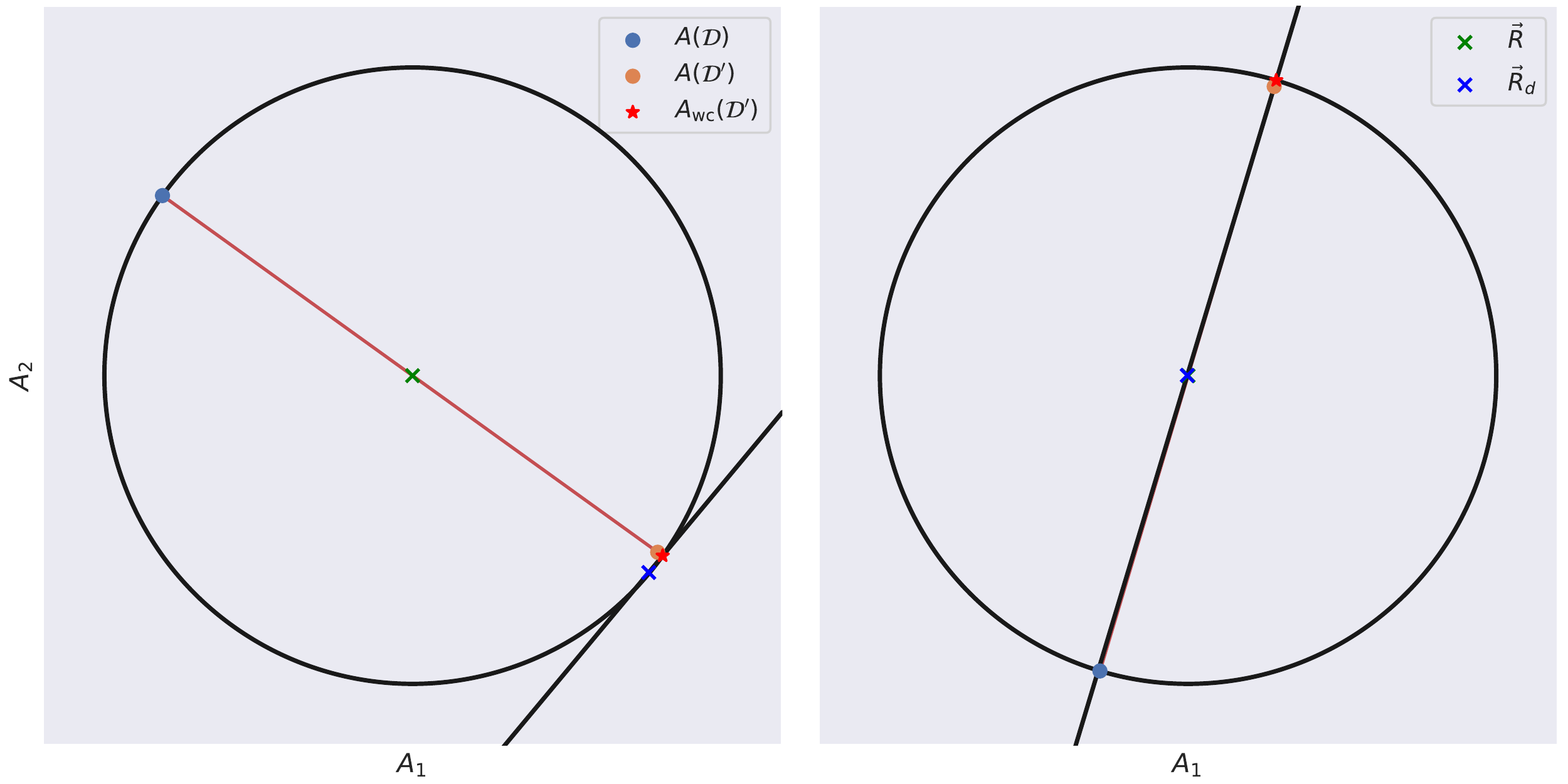}
    \caption{Pictorial demonstration of the relationship between the bounds from \citep{okumura2015quick} and \citep{zhou2020tighter} and the exact trained base and neighboring model points $A(x)$ and $A(y_i)$. The left-hand panel shows the example with the largest distance between sphere center and plane, and the right-hand panel shows the example with the largest $|A_{\mathrm{wc}}(y_i)-A(y_i)|$.}
    \label{fig:bounds}
\end{figure}

In Fig.~\ref{fig:bounds}, we demonstrate pictorially what the bounds look like for a couple of notable examples from the two-feature Adults dataset that we used throughout the main text. The left-hand panel shows the example with the largest $|A_{\mathrm{wc}}(y_i)-A(y_i)|$, and the right-hand panel shows the example with the smallest distance between sphere center and plane. The former implies that $A_{\mathrm{wc}}(y_i)$ is a good approximation to $A(y_i)$ in all cases, as suggested by Sec.~\ref{sec:bounds}. The latter shows the maximum ability for the bounds from Ref.~\citep{zhou2020tighter} to restrict the domain available to $A(y_i)$. We note that $A_{\mathrm{wc}}(y_i)$ lies within the bounds nonetheless.

As mentioned already in Sec.~\ref{sec:bounds}, we have empirically seen that the worst-case points defined here provide a very good approximation to the exact retrained model points on neighboring datasets and therefore provide an efficient method for computing the coverage. We note that this is dependent on the data normalization; we have seen that other normalization strategies do not give the same closeness of results.

\subsection{Coverage for Individual Model Points}
Using the formalism developed above, we compute coverage for a logistic regression model over the same 2-feature version of the Adults dataset used in Sec.~\ref{sec:motivation}. We remind the reader the purpose of the 2-feature version is to visually demonstrate the coverage effect, and our results also include the full-version of the Adult dataset. 
We use our method of computing the neighboring data point as described in \autoref{sec:bounds} in $\mathcal{O(D)}$ time without having to re-train on the entire dataset. 
In the Supplemental Material, we also replicate Fig.~\ref{fig:coverage} using this method together with Laplace output perturbation, emphasizing the visual similarity and qualitative agreement between the two sets of results. 

\begin{figure}[h!]
    \centering
    \begin{subfigure}[t]{0.48\textwidth}
    \centering
    \includegraphics[width=\textwidth]{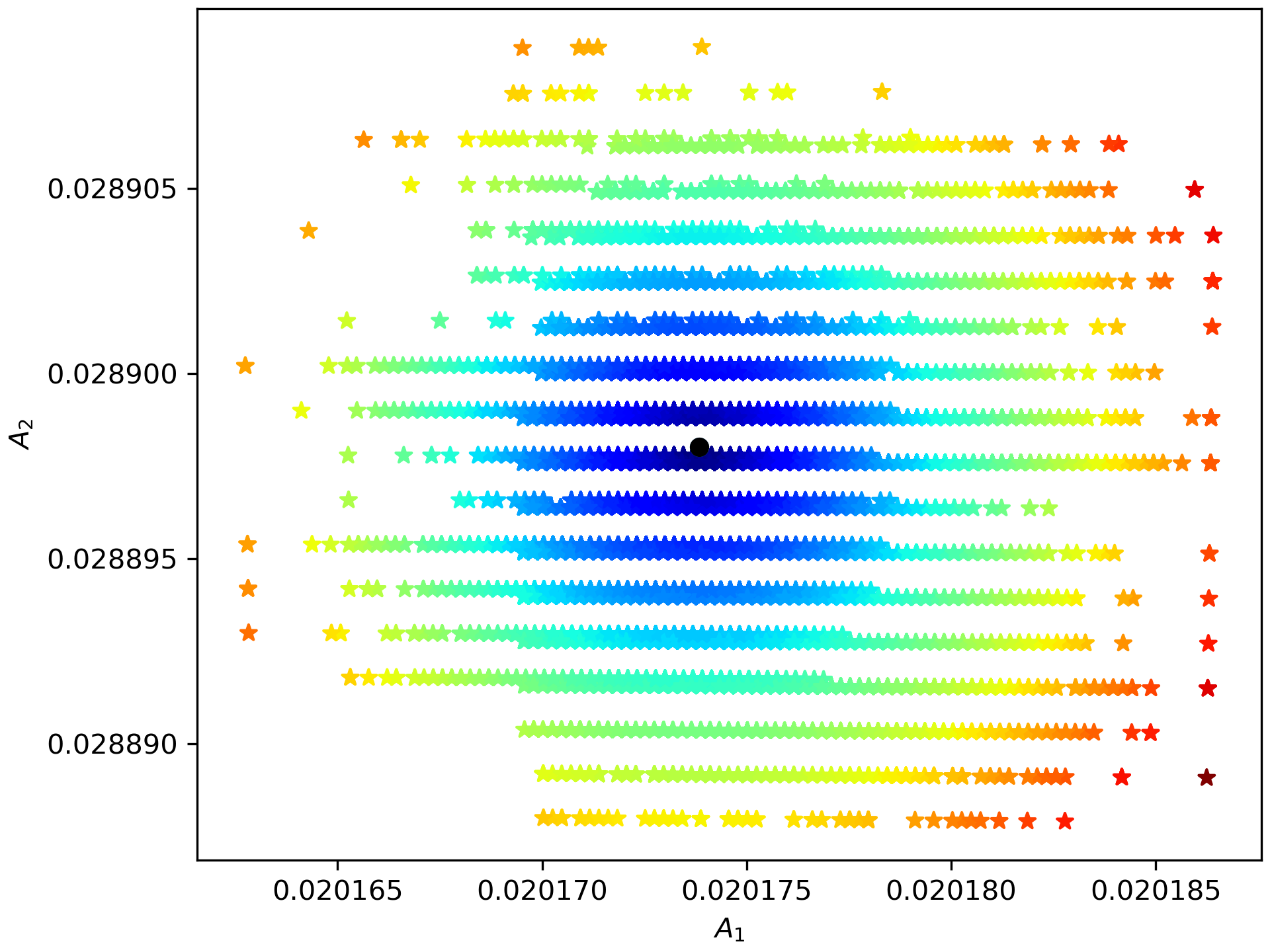}
    \caption{}
    \label{fig:coverage_heatmap1}
    \end{subfigure}
    \begin{subfigure}[t]{0.48\textwidth}
    \centering
    \includegraphics[width=\textwidth]{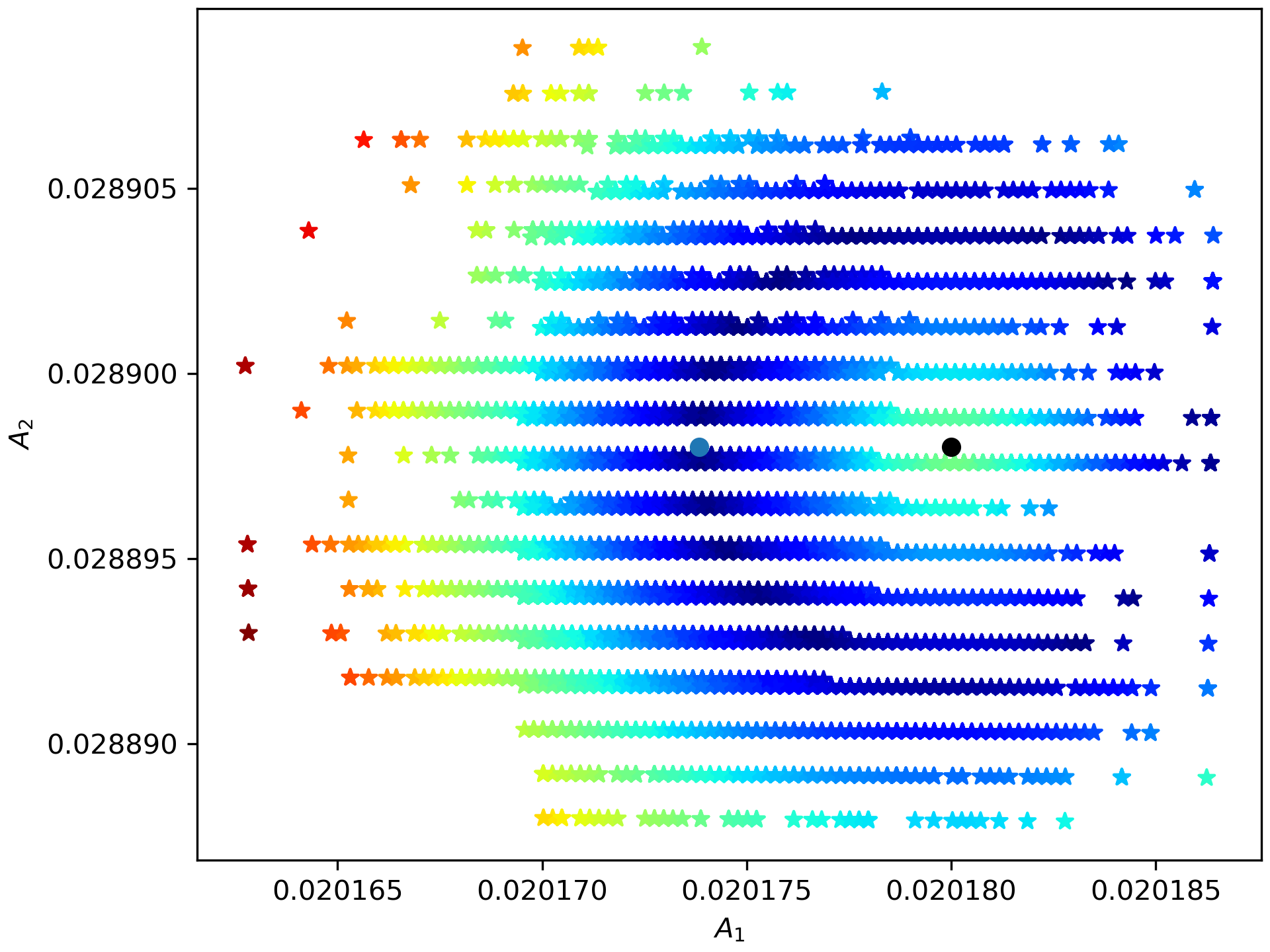}
    \caption{}
    \label{fig:coverage_heatmap2}
    \end{subfigure}
    \caption{Privacy loss ($|l(x,y_i,M)|$) heatmap for $M$ (a) centered at $A(x)$ and (b) off-center from $A(x)$. Stars represent the corresponding neighboring model point for each privacy loss value. Cooler colors indicate lower privacy losses. The dataset in question is the full (two-column) adults dataset.}
\end{figure}

To visualize the privacy profiles of two example models, Figures~\ref{fig:coverage_heatmap1} and~\ref{fig:coverage_heatmap2} are heatmaps that show the privacy losses for two $M$ (black circles). In the first, $M$ is placed exactly at $A(x)$. In the second, $M$ is placed off-center from $A(x)$, demonstrating the qualitative difference between these cases. In both figures, each of the stars, located at $A(y_i)$, corresponds to a different neighboring dataset. Cooler colors indicate a lower privacy loss and warmer colors indicate a higher one. In Fig.~\ref{fig:coverage_heatmap1}, one can see the privacy loss decaying monotonically away from the center, whereas in Fig.~\ref{fig:coverage_heatmap2}, the privacy loss is minimal for neighboring points for which $|A(y_i) - M| = |A(x) - M|$, as expected from Eq.~\eqref{eq:ploss3}. We envision the privacy profile being useful to the practitioner who wishes to know the relative vulnerability of training samples in their final model. 

\subsection{Coverage for Typical Model Points}

\begin{figure}[!h]
    \centering
    \begin{subfigure}[t]{0.48\textwidth}
    \centering
    \includegraphics[width=\textwidth]{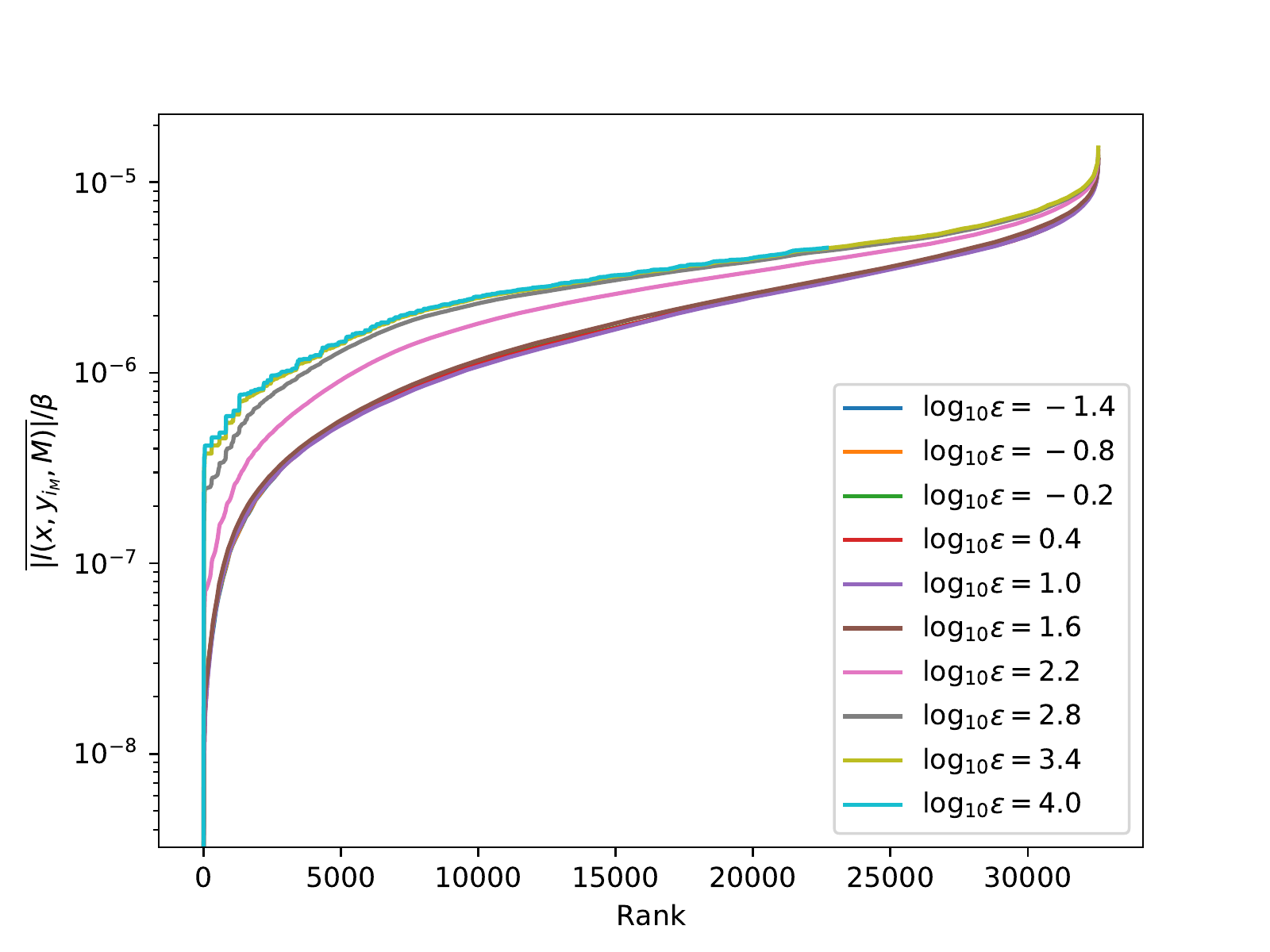}
    \caption{}
    \label{fig:rank}
    \end{subfigure}
    \begin{subfigure}[t]{0.48\textwidth}
    \centering
    \includegraphics[width=\textwidth]{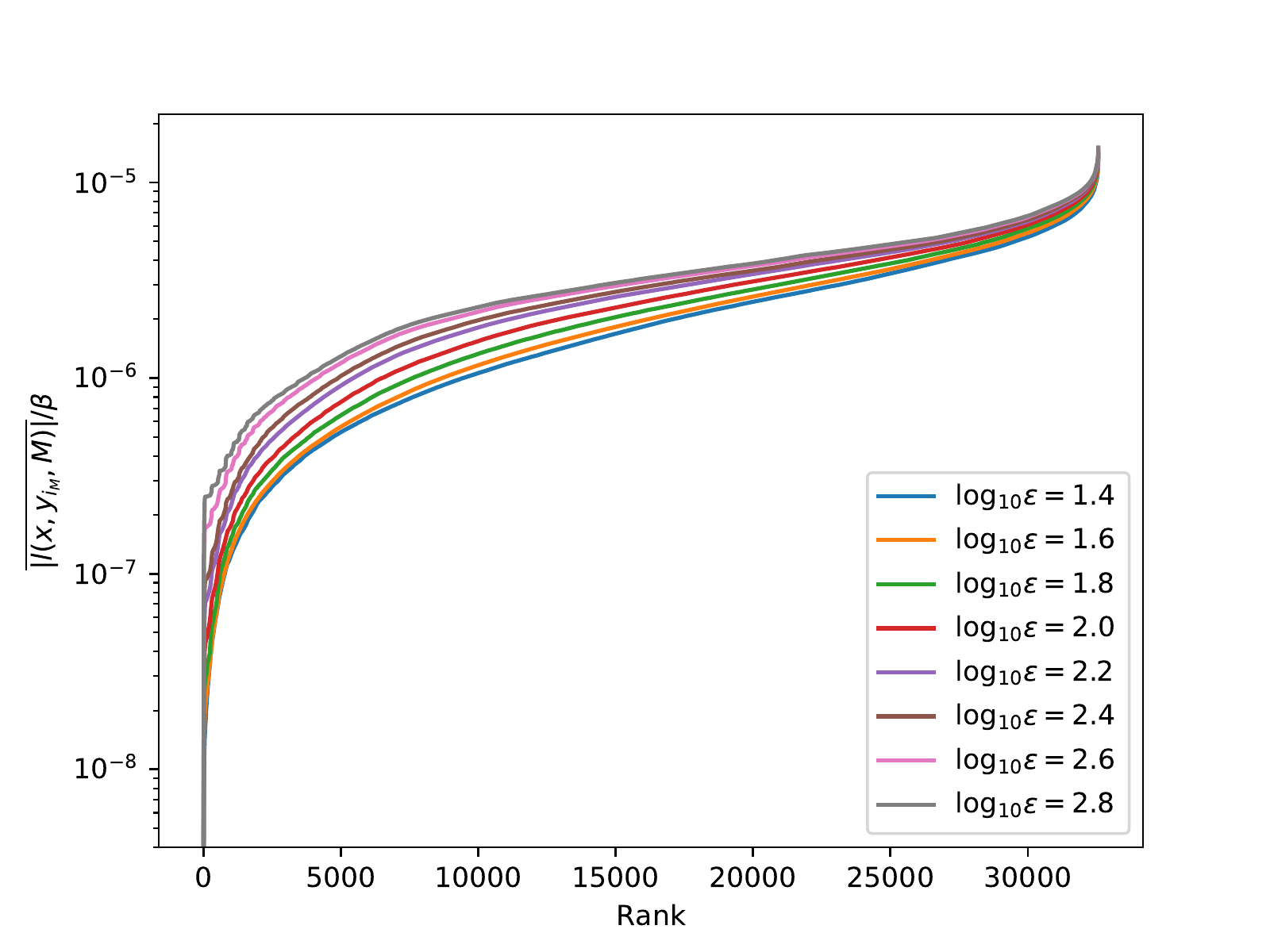}
    \caption{}
    \label{fig:rank2}
    \end{subfigure}
    \caption{Privacy profiles averaged over many $M$ sampled from $\mathcal{A}(x)$. The dataset in question is the full (two-column) adults dataset. Results are plotted for an array of $\epsilon$ values that (a) captures two stable regimes at high and low $\epsilon$ and (b) a transition regime for intermediate values of $\epsilon$. The practitioner can look for the bimodal  set of curves to determine which regime they are in. They want to find region (b) where their choice in $\epsilon$ has a meaningful impact on privacy. This also shows some data points will have high privacy (right highest rank) and others low privacy (left, near zero rank) over many ranges of $\epsilon$. This allows them to investigate sub-populations of the data that impact privacy as a whole and individual risk. }
\end{figure}

We next investigate the privacy profile for typical models in order to probe the transition to dataset-level privacy motivated in Sec.~\ref{sec:motivation}. In Fig.~\ref{fig:rank}, we plot $\overline{|l(x,y_{i_M},M)|}$ for different $\log_{10}\epsilon$, where $\overline{(...)}$ denotes an average over many samples of $M$. We normalize $\overline{|l(x,y_{i_M},M)|}$ by $\beta$ to reveal three scaling regimes. The first and third scaling regimes exist for high and low values of $\log_{10}\epsilon$, where it is clear that $\overline{|l(x,y_{i_M},M)|}\sim \beta$, indicating that $\overline{d(x,y_{i_M},M)}$ is invariant to $\beta$ in these regimes. The second regime, a transition between the low and high $\log_{10}\epsilon$ regimes in which $\overline{|l(x,y_{i_M},M)|}/\beta$ increases with $\epsilon$, is detailed further in Fig.~\ref{fig:rank2}. This narrow transition region coincides with the transition to distribution indistinguishabilty upon decreasing $\epsilon$ that was first noted in Section \ref{sec:motivation}. These results strongly suggest that the typical privacy profile defined here can be used by the ML practitioner as a powerful probe to determine a critical value of $\epsilon$ beneath which all rows of the training set are effectively protected for the typical model point sampled from an $\epsilon$-DP mechanism.

\begin{figure}[!h]
    \centering
    \includegraphics[width=\columnwidth]{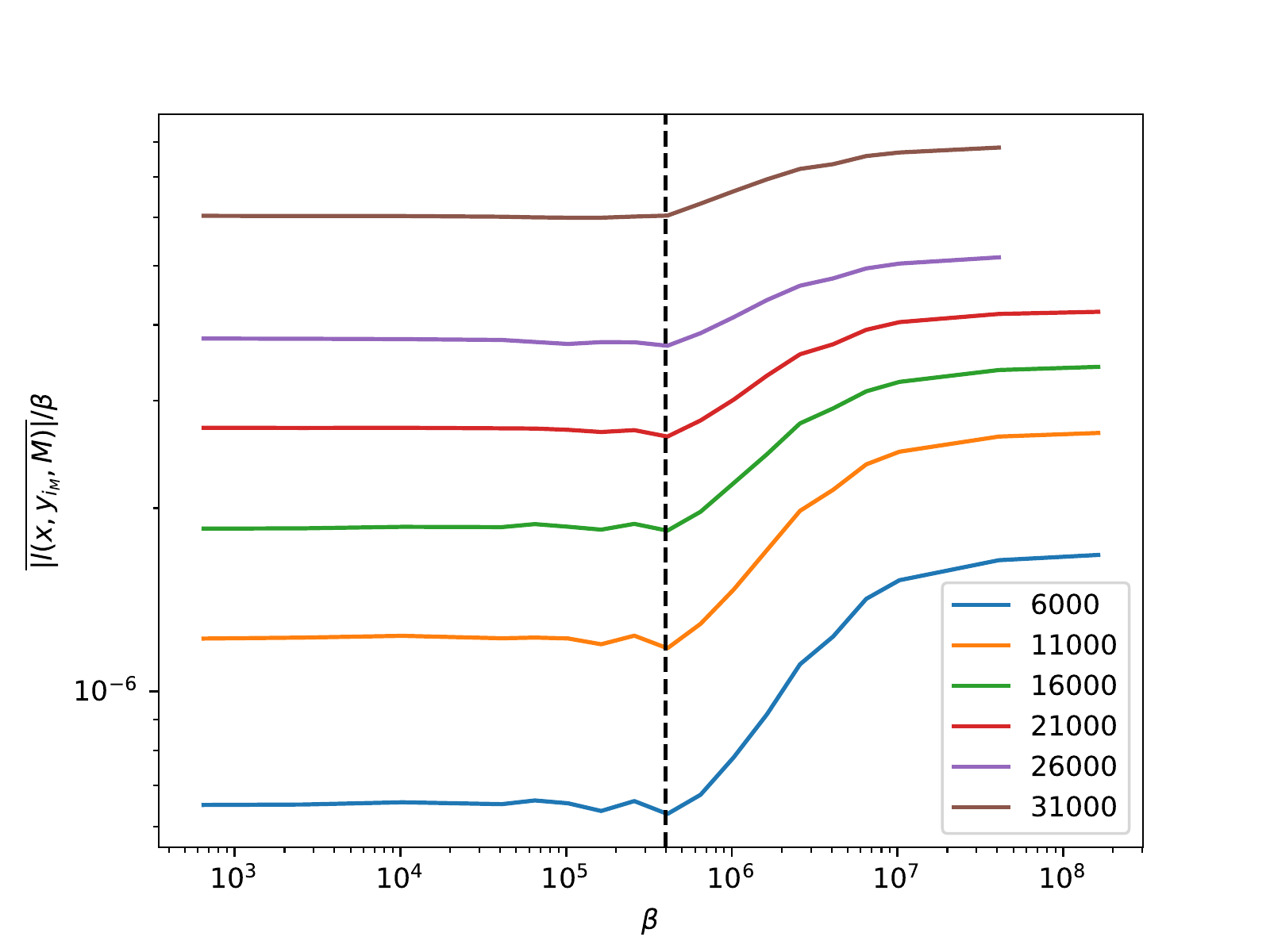}
    \caption{Values of the averaged privacy profiles from Figs.~\ref{fig:rank} and~\ref{fig:rank2} at different rank indices (see legend) are plotted against $\beta$ to demonstrate the transition between three regimes in $\beta$. This figure shows for individual data points privacy tends to start increasing (dashed line) at a shared point, and the amount of privacy per individual varies --- with increases plateauing. Note this reveals an order of magnitude range in which $\epsilon$ should be searched in practice, rather than a multiple order of magnitude range usually used in practice. }
    \label{fig:ploss_vs_eps2}
\end{figure}
\begin{figure*}
    \centering
    \includegraphics[width=0.99\textwidth]{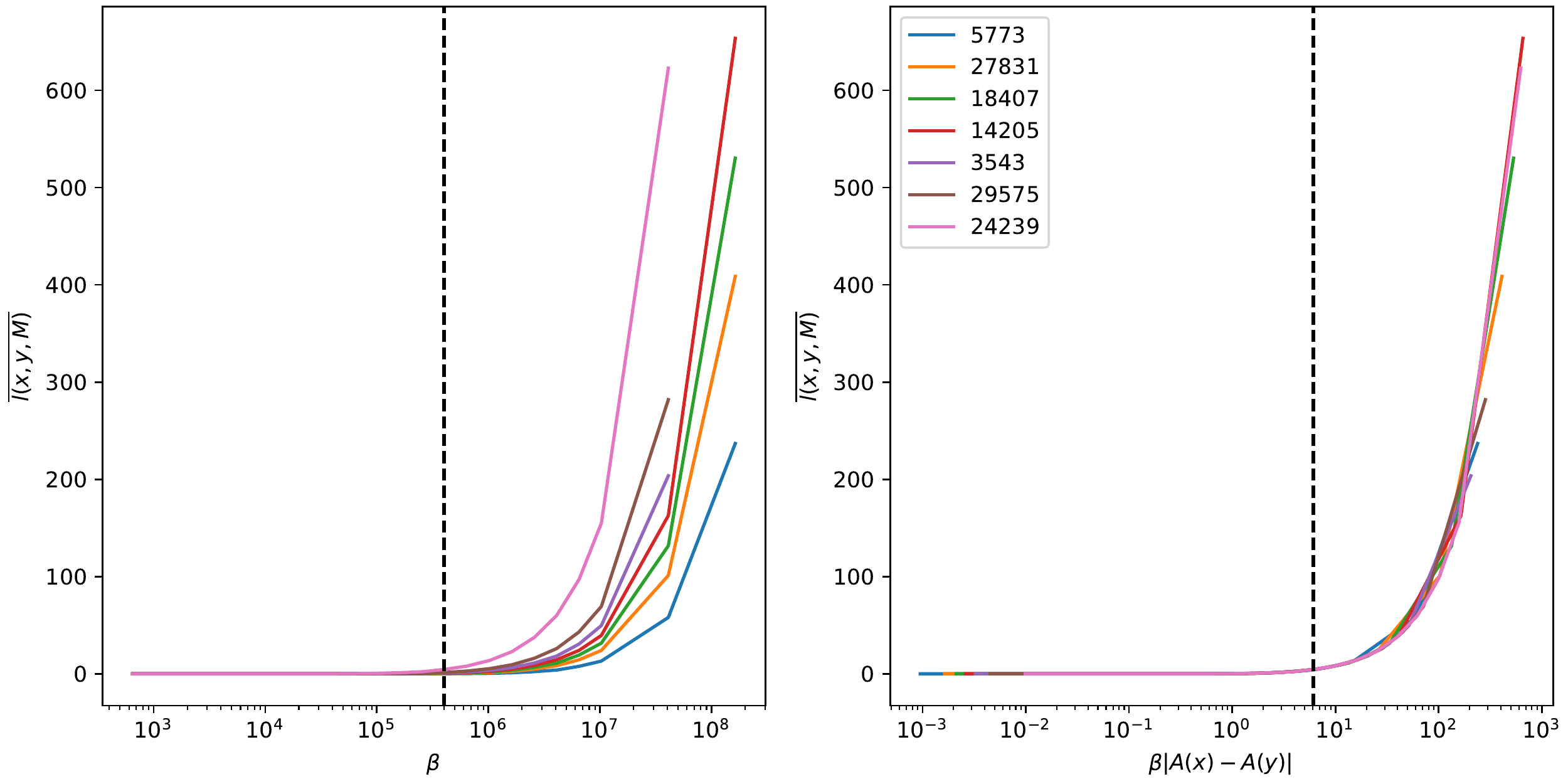}
    \caption{Average privacy loss $\overline{l(x,y_i,M)}$ vs. $\beta$ (left panel) and vs. $\beta |A(x)-A(y_i)|$ (right panel). Different curves correspond to different neighboring datasets $y_i$ (legend gives $i$) including $y_{\mathrm{max}}$ at $i=24239$. The black dotted line is the same as from Fig.~\ref{fig:ploss_vs_eps2} (scaled by $\beta |A(x)-A(y_{\mathrm{max}})|$ in right panel).}
    \label{fig:ploss_vs_eps_neighbors}
\end{figure*}

The onset values of the low and high $\epsilon$ regimes respectively correspond to points at which reducing $\epsilon$ gains no benefit in privacy for a typical model on the present dataset, and to which no more privacy is lost as $\epsilon$ increases. To better understand the behavior of $\overline{|l(x,y_{i_M},M)|}$ at the transition, in Fig.~\ref{fig:ploss_vs_eps2} we plot $\overline{|l(x,y_{i_M},M)|}/\beta$ vs. $\beta$ for different rankings across the spectrum. Each curve exhibits a plateau until $\beta$ reaches a critical value, at which point $\overline{|l(x,y_{i_M},M)|}/\beta$ increases until the onset of a second plateau. A dashed line is a guide to the eye showing the approximate onset of the transition regime from the initial plateau. The onset of the second plateau, on the other hand, appears index-dependent, with higher indices (toward greater privacy losses) having a narrower transition regime. The latter point is consistent with our intuition that close-by distributions will begin to appreciably cover sampled $M$ at higher values of $\beta$ than those which are farther away. According to these findings, beneath a particular $\beta$ cutoff and for typical samples $M$ from $\mathcal{A}(x)$, $d(x,y_{i_M},M)$ has converged to a form that is invariant to $\beta$. Beneath this $\beta$, any additional noise added only serves to send typical $M$ farther away from \textit{both} distribution centers $A(x)$ and $A(y_{i_M})$ at the same rate, thus decreasing accuracy without improving coverage.

Figure~\ref{fig:ploss_vs_eps_neighbors}, which shows privacy losses $\overline{l(x,y_i,M)}$ for various neighboring datasets $\{y_i\}$ throughout the spectrum (including $y_{\mathrm{max}}$), demonstrates that mean privacy losses plateau below the $\beta$ cutoff found in Fig.~\ref{fig:ploss_vs_eps2} (dashed black line), implying that typical privacy losses have reached a plateau for \textit{all} neighboring datasets beneath this threshold. In the right panel of Fig.~\ref{fig:ploss_vs_eps_neighbors}, we show that the privacy loss plateau onset scales inversely with $|A(x) - A(y_i)|$, implying that the $\beta^*$ needed to protect all rows of the dataset scales as $\beta^* \sim 1/|A(x)-A(y_{\mathrm{max}})|$, which was expected.

\section{Related Work} \label{sec:related_work}

There have been many approaches to contextualizing differential privacy. Membership inference attacks have been well-established as mechanisms for bounding differential privacy parameters \citep{yeom2018privacy}. Since then, others \citep{humphries2020datadeps, erlingsson2019secret} have tightened the bound. However, IID assumptions in data introduce unrealistic expectations for this as a measure \citep{humphries2020datadeps, tschantz2020causal}. Previous examinations \citep{yaghini2019bias, bagdasaryan2019disparate} emphasize that this assumption is impractical, showing that data bias consequences are passed to adversarial settings. These works are complementary to our own, which explicitly relate data dependencies to differentially private model behavior. Lower bounds for $\epsilon$ have been directly related to the model generalization gap \citep{he2020generalization}, though this requires model re-training regimes.

Numerous previous works provide approaches for the selection of DP parameters, namely $\epsilon$. \cite{hsu2014economicselection} offer an intuitive interpretation of the $\epsilon$ parameter as an economic event cost function within a multi-agent setting. \cite{ye2019miaselection} evaluate selection from within an MIA adversarial context across distributed privacy scenarios.
\cite{tsou2019accselection} provide model performance selection approaches. \cite{kohli2018votingselection} express $\epsilon$ optimization directly through user preference balancing. Coverage specifically examines the behavior of $\epsilon$ through realistic data dependencies with regard to the model parameter space.

Current approaches for quantifying the vulnerability of a sample are varied. Frameworks \citep{murakonda2020ml, art2018} compare randomly sampled attack profiles to order sample vulnerability. \citep{Lu2022AGF,dpsniper, ding2018detecting} exhaustively search the sample space for optimally poisoned samples, but do not extend tractably toward weighting neighboring datasets.

As mentioned, our approach is related to a variety of approaches to making DP more practical. \cite{dwork2016concentrated} have introduced the privacy loss random variable. \cite{cuff2016differential} have used the expected value of the privacy loss random variable in KL-DP. \cite{soria2017individual} have introduced individual DP, which is a dataset-constrained notion of privacy. \cite{wang2019per} has introduced per-instance DP (pDP), which quantifies the privacy of individual dataset samples. Finally, \cite{redberg2021privately} follow up on pDP by conditioning it to mechanism output in \textit{ex-post} pDP. This last work is the most closely related to our own and can be seen as complementary, as it presents efficient methods to calculate the same privacy loss used here, but in other contexts. 

\section{Conclusion} \label{sec:conclusion}

We have proposed the concept of coverage as a new lens through which we may apply DP to machine learning problems. Coverage probes the overlap of a given model's parameters with the DP training mechanism over neighboring datasets, allowing us to infer individual data point level risks based on the extent to which their removal incurs a specific privacy loss. We define a privacy profile to quantify row-level privacy in individual models and use it to probe an observed transition to indistinguishability of neighboring distributions in the output space.

Limitations of this work include the fixed feature space dimension $d$ and regularization parameter $\Lambda$ used throughout. Future work should emphasize the dependence of the typical privacy profile and $d(x, y_{i_M}, M)$ on these parameters, especially for $\beta$ beneath the transition to dataset-level privacy. Additionally, unpublished results have demonstrated that the onset of the plateau in typical privacy loss for any given neighbor is related to the location of a coverage peak; namely, we have observed that as the neighboring distributions expand with decreasing $\beta$, the typical $C(y_i, M)$ increases until it reaches a maximum value, at which point it decays in identical fashion to the decay of $C(x, M)$, begetting the observed plateau in typical $l(x, y_i, M)$. Future work should strengthen this intuition through analytical derivations or further numerical exploration. Lastly, it would be interesting to discover how these results extend to nonlinear models such as random forests, or to extend the analysis to $L_1$ regularized linear models ~\cite{Khanna2023SparsePL}.

\bibliography{main}

\clearpage
\appendix
\onecolumn

\section{Our Method Produces the Same Result}\label{sec:motivation2}
\begin{figure}[!h]
    \centering
    \includegraphics[width=\textwidth]{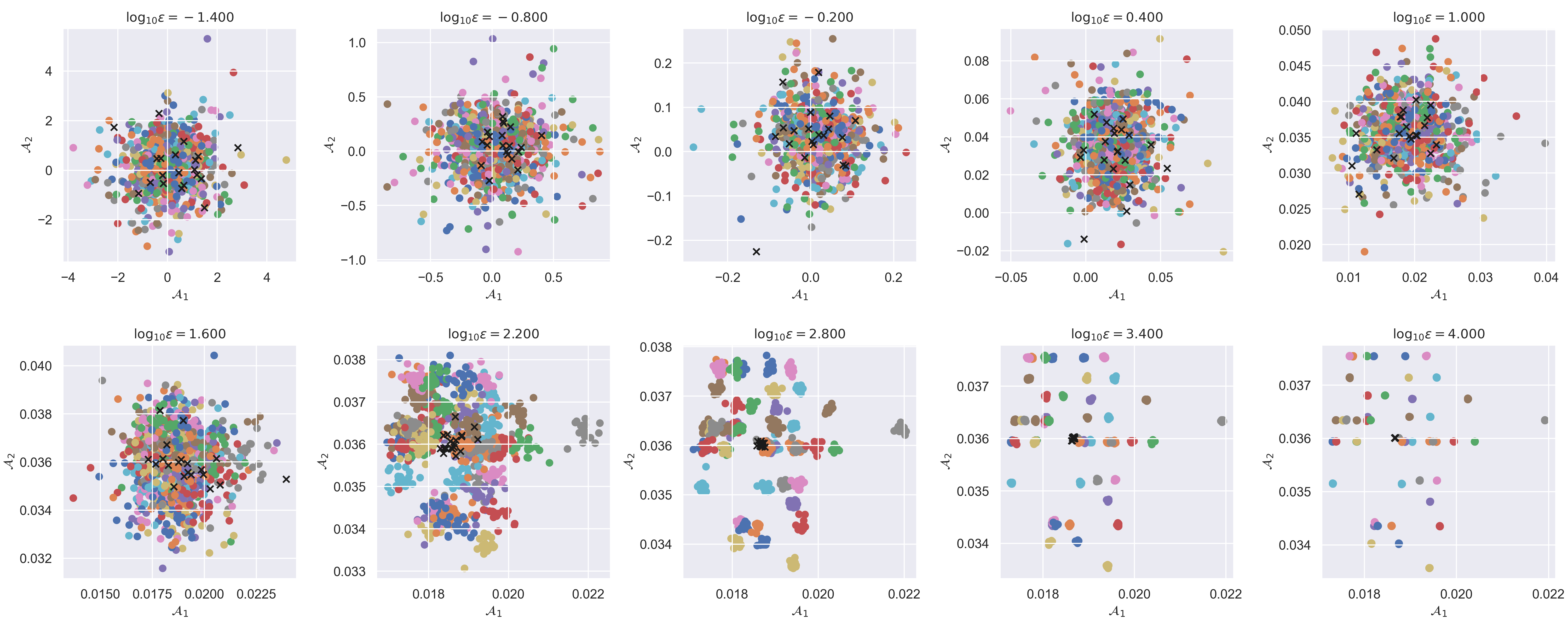}
    \caption{Model points sampled using output perturbation with subplot titles showing $\epsilon$ values. $A(y_i)$ were obtained from $A(x)$ for neighboring datasets using the process from Sec.~\ref{sec:bounds}. The dataset used is 100 rows of a 2-feature version of the Adults dataset (black x's). Colored circles are the results on 50 neighboring datasets. Compare with Fig.~\ref{fig:coverage} to see closeness of our results to samples from Diffprivlib.}
    \label{fig:coverage2}
\end{figure}

In Fig.~\ref{fig:coverage2}, we have reproduced Fig.~\ref{fig:coverage} through the computation of neighboring model points using the method from Sec.~\ref{sec:bounds} and implementation of $\epsilon$-DP via the Laplace output perturbation method from \cite{chaudhuri2011differentially}. One can see that Fig.~\ref{fig:coverage2} is visually similar to Fig.~\ref{fig:coverage}, and that all of the observations from Sec.~\ref{sec:motivation} equally apply. 

\end{document}